\documentclass[sigconf]{acmart}
\usepackage{graphicx}
\usepackage[most]{tcolorbox} 
\usepackage{amsmath}
\usepackage{mathtools} 
\usepackage{wrapfig}
\usepackage{algorithm}
\usepackage{booktabs}
\usepackage{multirow}
\usepackage{caption}   
\usepackage{placeins}
\usepackage[table]{xcolor}
\usepackage{algpseudocode}
\usepackage{subcaption}

\usepackage{geometry}
\usepackage{color}
\usepackage{pifont}
\newcommand{\cmark}{\ding{51}} 
\newcommand{\xmark}{\ding{55}} 
\newcommand{\boxedlink}[2]{%
  \href{#2}{%
    \begingroup
    \setlength{\fboxsep}{0.6pt}
    \setlength{\fboxrule}{0.3pt}
    \fbox{#1}%
    \endgroup
  }%
}
\AtBeginDocument{%
  }

\setcopyright{acmlicensed}
\copyrightyear{2018}
\acmYear{2018}
\acmDOI{XXXXXXX.XXXXXXX}

\acmConference[Conference acronym 'XX]{Make sure to enter the correct
  conference title from your rights confirmation emai}{June 03--05,
  2018}{Woodstock, NY}
\acmISBN{978-1-4503-XXXX-X/18/06}

\begin{document}

\title{HouseTS: A Large-Scale, Multimodal Spatiotemporal U.S. Housing Dataset and Benchmark}


\author{Shengkun Wang}
\affiliation{%
  \institution{Virginia Tech}
  \city{Alexandria}
  \state{Virginia}
  \country{USA}
}

\author{Yanshen Sun}
\affiliation{%
  \institution{Virginia Tech}
  \city{Alexandria}
  \state{Virginia}
  \country{USA}
}

\author{Fanglan Chen}
\affiliation{%
  \institution{Virginia Tech}
  \city{Alexandria}
  \state{Virginia}
  \country{USA}
}

\author{Linhan Wang}
\affiliation{%
  \institution{Virginia Tech}
  \city{Alexandria}
  \state{Virginia}
  \country{USA}
}

\author{Naren Ramakrishnan}
\affiliation{%
  \institution{Virginia Tech}
  \city{Alexandria}
  \state{Virginia}
  \country{USA}
}

\author{Chang-Tien Lu}
\affiliation{%
  \institution{Virginia Tech}
  \city{Alexandria}
  \state{Virginia}
  \country{USA}
}

\author{Yinlin Chen}
\affiliation{%
  \institution{Virginia Tech}
  \city{Blacksburg}
  \state{Virginia}
  \country{USA}
}


\begin{abstract}


Accurate long-horizon house-price forecasting requires benchmarks that capture temporal dynamics together with time-varying local context. However, existing public resources remain fragmented: many datasets have limited spatial coverage, temporal depth, or multimodal alignment; the robustness of modern deep forecasters and time-series foundation models on housing data is not well characterized; and aerial imagery is rarely leveraged in a time-aware and interpretable manner at scale. To bridge these gaps, we present \textbf{HouseTS} (\textbf{House} \textbf{T}ime \textbf{S}eries), a multimodal spatiotemporal dataset for ZIP-code-level housing-market analysis, covering monthly signals from March 2012 to December 2023 across over 6{,}000 ZIP codes in 30 major U.S.\ metropolitan areas. HouseTS aligns monthly housing-market indicators, monthly POI dynamics, and annual census-based socioeconomic variables under a unified schema, and includes time-stamped annual aerial imagery. Building on HouseTS, we define standardized long-horizon forecasting tasks for univariate and multivariate prediction and benchmark 16 model families spanning statistical methods, classical machine learning, deep neural networks, and time-series foundation models in both zero-shot and fine-tuned modes. We also provide image-derived textual change annotations from multi-year aerial image sequences via a vision--language pipeline with LLM-as-judge and human verification to support scalable interpretability analyses. HouseTS is available on \boxedlink{Kaggle}{https://www.kaggle.com/datasets/shengkunwang/housets-dataset}, with code and documentation on \boxedlink{GitHub}{https://github.com/hao1zhao/Housets_data_bench}.

\end{abstract}



\keywords{Spatiotemporal dataset, Housing prices, Time-series forecasting, Image-derived text annotations}

\maketitle

\section{INTRODUCTION}
\begin{figure*}[t]
    \centering
    \includegraphics[width=\textwidth]{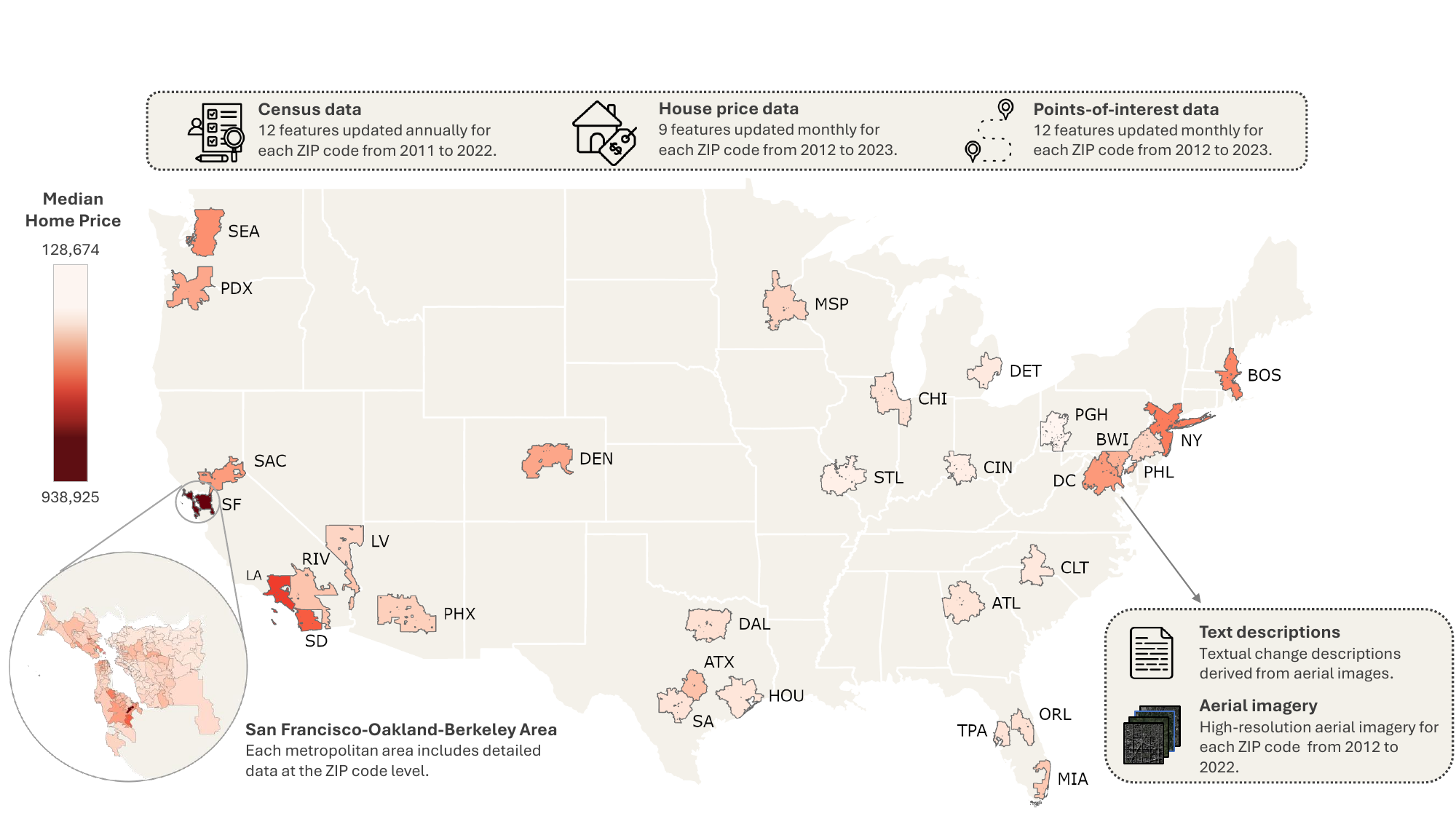}
    \caption{HouseTS overview: 30 U.S.\ metropolitan areas and aligned ZIP-level modalities.}
    \label{fig:main}
\end{figure*}

Accurate house-price prediction is vital for investors, policy makers, and researchers. However, most existing studies rely on narrow data sources, such as past sales or basic demographic counts, and often focus on individual properties without considering broader spatial and temporal patterns~\cite{8882834,truong2020housing,adetunji2022house,lu2023disamenity,park2015using,ho2021predicting}. Some recent works introduce multimodal inputs like satellite imagery or points of interest, but typically treat them as static features and fail to capture long-term dynamics~\cite{law2019take,kang2021understanding}. While survey papers provide useful overviews of data and methods~\cite{geerts2023survey,zulkifley2020house}, few offer an open and standardized dataset that reflects the full complexity of housing markets, including both physical and socioeconomic contexts. In addition, many existing time-series datasets in this domain are either too coarse in granularity or overly focused on high-frequency signals, making them unsuitable for long-term, regional housing analysis~\cite{shi2024timemoe}. This lack of comprehensive, well-aligned, and multimodal resources limits the development, evaluation, and interpretability of forecasting models. A dataset that combines long-term temporal coverage, rich contextual variables, and consistent preprocessing across diverse geographies is therefore urgently needed.

While data limitations pose one major challenge, modeling approaches in house price prediction also face constraints. Many existing studies still rely on statistical techniques or conventional machine learning models~\cite{ghourchian2024housing,soltani2022housing,rico2021machine,zhang2021spatial}. In parallel, the broader time-series forecasting community has made significant progress with deep architectures, including Transformer-based models and large pretrained models designed for sequential data. However, recent research has raised concerns about their robustness and generalizability. Although these models perform well on standard benchmarks such as M4~\cite{makridakis2020m4}, their accuracy often declines on domain-specific or real-world datasets~\cite{tan2024language,zeng2023transformers,bergmeir2024llms,godahewa2021monash}. To better understand their effectiveness in the housing context, we use our newly curated dataset to rigorously evaluate and compare deep neural networks and time-series foundation models against traditional baselines, with a focus on long-term, multivariate house price prediction.

Beyond forecasting accuracy, housing analysis also calls for interpretable, time-aware signals that help explain local market dynamics. Aerial imagery is a promising source of such context, but even when it is available, it is rarely used in an interpretable and temporally grounded way at scale~\cite{xia2015accurate}. Raw images are high-dimensional and noisy, and typical pipelines either depend on task-specific feature engineering or end-to-end vision models that are difficult to audit, sensitive to missing years and heterogeneous image quality, and costly to curate with human labels~\cite{xia2015accurate,southworth2024machine}. As a result, the visual modality is often reduced to static embeddings or single-year snapshots, providing limited insight into how neighborhood form evolves and which visible changes plausibly relate to downstream housing-market dynamics~\cite{bency2017beyond}.

To address these challenges, we introduce \textbf{HouseTS}, an aligned multimodal spatiotemporal dataset for ZIP-code-level housing-market analysis. HouseTS integrates long-term housing-market signals with dynamic local context and time-stamped aerial imagery, and supports standardized long-horizon forecasting benchmarks and interpretable multimodal analyses. Our contributions are threefold:

\textbf{Dataset:} We release \textbf{HouseTS}, covering over 6{,}000 ZIP codes across 30 major U.S.\ metropolitan areas from 2012 to 2023. It aligns monthly housing-market indicators, monthly neighborhood amenity dynamics from POIs, and annual census-based socioeconomic variables under a unified schema, and we also provide the raw, unimputed source tables.

\textbf{Benchmark:} We define standardized long-horizon forecasting tasks for univariate and multivariate prediction with consistent evaluation protocols, and benchmark 16 model families spanning statistical methods, classical machine learning, deep neural networks, and time-series foundation models in both zero-shot and fine-tuned modes; our results show that strong preprocessing is critical for neural stability and that simple regularized linear baselines remain highly competitive across horizons.

\textbf{Interpretability:} We provide time-stamped high-resolution annual aerial imagery, along with imagery-based interpretability resources. These include image-derived textual change annotations produced by a vision--language pipeline with an LLM-as-judge step and human verification, enabling time-aware and human-readable exploratory analyses of neighborhood change alongside the tabular benchmark.

\section{Related Work}

\renewcommand{\cmark}{\textcolor{green!60!black}{\ding{51}}}
\renewcommand{\xmark}{\textcolor{red}{\ding{55}}}

\begin{table*}[t]
\centering
\small
\resizebox{\textwidth}{!}{%
\begin{tabular}{@{}l c c c|c l l l l l@{}}
\toprule
\textbf{Data source \& Research paper} &
\textbf{Tabular} &
\textbf{Image} &
\textbf{Text} &
\textbf{Time stamp} &
\textbf{Frequency} &
\textbf{Time span} &
\textbf{Spatial unit} &
\textbf{Observations} &
\textbf{Model types}\\ \midrule
\href{https://www.kaggle.com/datasets/harlfoxem/housesalesprediction/data}{House Sales in King County}~\cite{8882834} & \cmark & \xmark & \xmark & \cmark & Daily & 1 year & Property & 21.6K & Stat,ML \\
\href{https://www.kaggle.com/datasets/ruiqurm/lianjia}{Housing Price in Beijing}~\cite{truong2020housing} & \cmark & \xmark & \xmark & \cmark & Daily & 8 years & Property & 319K & Stat \\
\href{https://www.kaggle.com/datasets/altavish/boston-housing-dataset}{Boston Housing Dataset}~\cite{adetunji2022house} & \cmark & \xmark & \xmark & - & - & - & Property & 0.5K & Stat \\
\href{https://insideairbnb.com/get-the-data/}{Airbnb}~\cite{camatti2024predicting} & \cmark & \xmark & \cmark & - & - & - & Property & 142K & Stat,DNN \\
\href{https://planet.openstreetmap.org/}{OpenStreetMap}~\cite{kang2021understanding} & \cmark & \cmark & \xmark & - & - & - & Property & 470K & ML \\
\href{https://mapsplatform.google.com/}{Google Map}~\cite{law2019take} & \cmark & \cmark & \xmark & - & - & - & Region & 111K & DNN \\
\href{https://www.dallasfed.org/research/international/houseprice\#data}{International House Price Database}~\cite{wang2023can} & \cmark & \xmark & \xmark & \cmark & Quarterly & 46 years & Country & 4.4K & Stat \\
\href{https://www.fhfa.gov/data/hpi/datasets?tab=quarterly-data}{FHFA HPI}~\cite{malliaris2024one} & \cmark & \xmark & \xmark & \cmark & Quarterly & 49 years & State & 9.3K & Stat \\
\href{https://www.redfin.com/news/data-center/}{Redfin}~\cite{kang2021understanding} & \cmark & \xmark & \xmark & - & - & - & Property & 125K & ML \\
\href{https://www.zillow.com/research/data/}{Zillow}~\cite{lu2023disamenity} & \cmark & \xmark & \xmark & \cmark & Daily & 5 years & Property & 1905K & Stat \\
\rowcolor{pink!30}
\textbf{HouseTS (Ours)} & \cmark & \cmark & \cmark$^\dagger$ & \cmark & Monthly & 11 years & Region & 890K & Stat,ML,DNN,Foundation \\
\bottomrule
\end{tabular}}
\caption{Comparison of HouseTS with existing housing datasets and related studies. Left: data source modalities, indicating the types of raw inputs available. Right: how these datasets have been used in past research, including temporal setup, spatial scope, data scale, and model types. $^{\dagger}$Text in HouseTS is derived in VLM-generated descriptions.}
\label{tab:related_work}
\end{table*}

House price modeling draws on heterogeneous signals, ranging from macro socioeconomic indicators to micro-level transaction and property attributes and neighborhood amenities~\cite{ghourchian2024housing,mian2011house,park2015using,ho2021predicting}.  While these methods offer useful insights, they are often limited by narrow geographic scope and short time spans, making them inadequate for modeling long-term trends and spatial variation. More data types, such as satellite imagery~\cite{law2019take,wang2021deep}, environmental conditions~\cite{xiao2017exploring}, and points of interest~\cite{yang2018walking}, offer richer contextual information for prediction. However, these sources differ in spatial resolution, update frequency, and structure, which makes them difficult to integrate into a unified modeling framework.

\textbf{A variety of open-source datasets} have been proposed for house price research, offering either property-level features such as physical attributes, transaction histories, and neighborhood amenities~\cite{shree2018house,becker2018melbourne,nugent2017california,pino2017melbourne}. or aggregated indices for broader market trends~\cite{FHFA_HPI,Zillow_ZHVI,Redfin_Data,Realtor_Data}. Building upon these resources, house price forecasting has emerged as an active research area, employing a range of techniques at both the individual property level~\cite{park2015using,ho2021predicting} and the regional level~\cite{ghourchian2024housing,mian2011house}. Diverse sources of information have been explored, including real estate transaction records~\cite{de2011ames}, textual descriptions~\cite{xu2021new}, environmental conditions~\cite{xiao2017exploring}, satellite imagery~\cite{wang2021deep}, census statistics~\cite{montero2018housing}, and points of interest data~\cite{yang2018walking} have been explored to enrich modeling capabilities. However, most existing datasets focus on a single city or cover only short time spans, limiting the potential for long-horizon analyses and cross-region comparisons. 

\textbf{Housing price models} have not been systematically evaluated under recent time-series forecasting advances. While Transformer-based and pretrained models perform strongly on standard benchmarks, prior work reports limited robustness and transfer to domain-specific datasets~\cite{tan2024language,zeng2023transformers,bergmeir2024llms,godahewa2021monash}.
How these models behave on housing markets, with strong spatial heterogeneity and socioeconomic drivers, is therefore less clear. 
Using HouseTS, we benchmark a diverse set of methods, covering classical statistical baselines, traditional machine learning models, deep neural networks, and pretrained time-series foundation models: 

\emph{Statistical approaches and classical machine learning methods:}
Statistical models such as AR~\cite{box2015time} and ARDL~\cite{pesaran2001bounds}, as well as traditional machine learning algorithms like Random Forests~\cite{breiman2001random} and XGBoost~\cite{chen2016xgboost}, have been widely used in house price forecasting. These methods are often favored for their interpretability, ease of implementation, and relatively low computational cost. 

\emph{Deep learning models:}
Deep learning methods have become standard baselines for time-series forecasting by modeling nonlinear temporal dependencies. Early approaches rely on recurrent architectures such as RNN and LSTM~\cite{elman1990finding,hochreiter1997long}, while more recent lightweight designs such as DLinear and TimeMixer use simple feed-forward components for efficient forecasting~\cite{zeng2023transformers,wang2023timemixer}. Transformer-based forecasters, including Informer, Autoformer, FEDformer, and PatchTST, further improve long-range modeling through attention-based sequence representations and structural inductive biases~\cite{zhou2021informer,wu2021autoformer,zhou2022fedformer,Yuqietal-2023-PatchTST}. For spatially linked time series, spatiotemporal graph neural networks model temporal dynamics together with graph-structured interactions. Representative baselines include STGCN and Graph WaveNet~\cite{yu2017spatio,wu2019graph}.

\emph{Pretrained time-series foundation model:}
Recent work has proposed large pretrained models for time-series forecasting, including Chronos~\cite{ansari2024chronos} and TimesFM~\cite{das2024decoder}. These models are trained on broad collections of time-series data and support zero-shot or few-shot forecasting across different domains. In principle, they offer strong generalization and can incorporate heterogeneous signals such as prices, economic indicators, and even text descriptions. Their modular design enables rapid adaptation without task-specific architectures. 

\section{HouseTS Dataset}
\begin{figure*}[t]
    \centering
    \includegraphics[width=\textwidth]{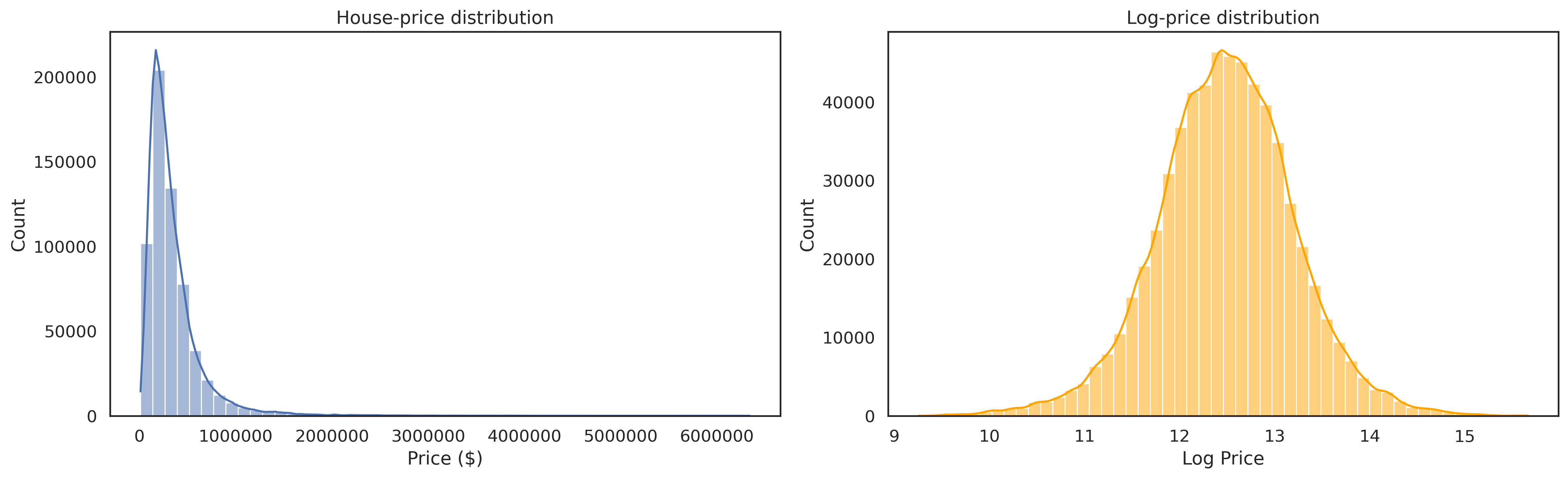}
    \caption{Distribution of house prices before (left) and after log transformation (right).}
    \label{price}
\end{figure*}

This section describes the construction of the \textbf{HouseTS} dataset, including data sources, spatiotemporal alignment, preprocessing, and basic analyses. We release both raw and cleaned versions of the data, together with preprocessing code and visualization notebooks. HouseTS is a ZIP-code-level multimodal panel covering 6{,}000 ZIP codes across 30 major U.S.\ metropolitan areas from March 2012 to December 2023, integrating housing-market signals, socioeconomic indicators, points of interest, and time-stamped aerial imagery. While long-horizon house-price forecasting is the primary benchmark reported in this paper, the aligned design supports broader spatiotemporal learning and multimodal analyses, including multivariate forecasting beyond the price series, structured-missingness imputation and denoising across modalities, cross-city transfer evaluation on held-out regions, and interpretability-oriented change understanding from imagery. Summary statistics are reported in Appendix Table~\ref{tab:desc_stats}.

\textbf{Dataset schema and spatiotemporal alignment.}
HouseTS is organized as a ZIP-code panel with a shared monthly timeline for all tabular modalities. Each tabular record corresponds to one ZIP code and one calendar month, and all tabular sources are aligned to this common index. Annual socioeconomic indicators from the American Community Survey are incorporated as yearly features and aligned to the monthly panel under a temporally valid rule that avoids leakage. Aerial imagery is provided as a separate auxiliary modality indexed by ZIP code and year.

\textbf{Historical house price-related features.}
HouseTS aggregates ZIP code level housing market signals from two widely used real-estate platforms, Zillow and Redfin. We focus on the all-residential category to maintain consistent definitions across metropolitan areas. From Zillow, we use the Zillow Home Value Index as the primary price signal. It provides a smoothed and seasonally adjusted estimate of the median home value at monthly resolution and serves as the main target in our forecasting benchmark~\cite{ZillowResearchData}. Redfin complements this target with a broader set of monthly market indicators at the ZIP code level~\cite{RedfinDataCenter}. These variables capture both pricing and market activity, including median sale and list prices, price per square foot, transaction volume, pending sales, new listings, inventory, and time-on-market measures. They also include liquidity and bargaining signals such as the sale-to-list ratio, the share sold above list, and the share off market within two weeks. We exclude derived growth-rate columns to reduce the propagation of errors under missing-value handling. Appendix Figure~\ref{fig:4} further visualizes cross-city heterogeneity in the price distributions.

{\setlength{\dblfloatsep}{-4pt}
 \setlength{\dbltextfloatsep}{2pt}
 \setlength{\abovecaptionskip}{2pt}
 \setlength{\belowcaptionskip}{0pt}

\begin{table*}[t]
\centering
\resizebox{\textwidth}{!}{%
\begin{tabular}{lcccccccccccc}
\toprule
\textbf{Window size $\rightarrow$} & \multicolumn{2}{c}{\{6,3\}} & \multicolumn{2}{c}{\{6,6\}} & \multicolumn{2}{c}{\{6,12\}} & \multicolumn{2}{c}{\{12,3\}} & \multicolumn{2}{c}{\{12,6\}} & \multicolumn{2}{c}{\{12,12\}} \\
\cmidrule(lr){2-3}\cmidrule(lr){4-5}\cmidrule(lr){6-7}\cmidrule(lr){8-9}\cmidrule(lr){10-11}\cmidrule(lr){12-13}
\textbf{Model $\downarrow$} & Log\textendash RMSE & MAPE & Log\textendash RMSE & MAPE & Log\textendash RMSE & MAPE & Log\textendash RMSE & MAPE &  Log\textendash RMSE & MAPE & Log\textendash RMSE & MAPE \\
\midrule

ARDL & \colorbox{pink!30}{0.0119} & \colorbox{pink!30}{0.0079} & \colorbox{pink!30}{0.0267} & \colorbox{pink!30}{0.0181} & \colorbox{red!30}{0.0516} & \colorbox{red!30}{0.0369} & \colorbox{pink!30}{0.0117} & \colorbox{pink!30}{0.0079} & \colorbox{pink!30}{0.0262} & \colorbox{pink!30}{0.0179} & \colorbox{red!30}{0.0498} & \colorbox{pink!30}{0.0353} \\

RandomForest & 0.0202 & 0.0138 & 0.0364 & 0.0253 & 0.0582 & 0.0439 & 0.0211 & 0.0142 & 0.0356 & 0.0256 & 0.0582 & 0.0437 \\

XGBoost & 0.0286 & 0.0173 & 0.0413 & 0.0269 & 0.0576 & 0.0424 & 0.0286 & 0.0171 & 0.0445 & 0.0283 & 0.0568 & 0.0412 \\

\cmidrule(lr){1-1}\cmidrule(lr){2-3}\cmidrule(lr){4-5}\cmidrule(lr){6-7}\cmidrule(lr){8-9}\cmidrule(lr){10-11}\cmidrule(lr){12-13}

RNN & 0.0259 & 0.0171 & 0.0329 & 0.0232 & 0.0625 & 0.0453 & 0.0305 & 0.0225 & 0.0401 & 0.0292 & 0.0669 & 0.0511 \\

LSTM & 0.0209 & 0.0158 & 0.0379 & 0.0290 & 0.0550 & 0.0411 & 0.0172 & 0.0120 & 0.0293 & 0.0208 & \colorbox{pink!30}{0.0533} & 0.0399 \\

STGCN & 0.0516 & 0.0413 & 0.0653 & 0.0525 & 0.1165 & 0.1069 & 0.0543 & 0.0439 & 0.0622 & 0.0509 & 0.1038 & 0.0810 \\

Graph WaveNet & 0.0302 & 0.0239 & 0.0404 & 0.0303 & 0.0762 & 0.0604 & 0.0309 & 0.0245 & 0.0505 & 0.0416 & 0.0635 & 0.0503 \\

DLinear & \colorbox{red!30}{0.0108} & \colorbox{red!30}{0.0070} & \colorbox{red!30}{0.0260} & \colorbox{red!30}{0.0173} & \colorbox{pink!30}{0.0520} & \colorbox{pink!30}{0.0370} & \colorbox{red!30}{0.0106} & \colorbox{red!30}{0.0067} & \colorbox{red!30}{0.0255} & \colorbox{red!30}{0.0168} & \colorbox{red!30}{0.0498} & \colorbox{red!30}{0.0350} \\

Autoformer & 0.0248 & 0.0194 & 0.0372 & 0.0275 & 0.0584 & 0.0443 & 0.0326 & 0.0282 & 0.0395 & 0.0305 & 0.0583 & 0.0446 \\

PatchTST & 0.0360 & 0.0280 & 0.0508 & 0.0410 & 0.0660 & 0.0519 & 0.0328 & 0.0249 & 0.0465 & 0.0357 & 0.0595 & 0.0452 \\

FEDformer & 0.0316 & 0.0271 & 0.0379 & 0.0284 & 0.0605 & 0.0470 & 0.0235 & 0.0176 & 0.0516 & 0.0452 & 0.0598 & 0.0461 \\

Informer & 0.0435 & 0.0374 & 0.0524 & 0.0425 & 0.0770 & 0.0616 & 0.0376 & 0.0288 & 0.0548 & 0.0453 & 0.0669 & 0.0524 \\

TimeMixer & 0.0142 & 0.0106 & 0.0312 & 0.0223 & 0.0651 & 0.0524 & 0.0139 & 0.0096 & 0.0354 & 0.0274 & 0.0561 & 0.0432 \\

\cmidrule(lr){1-1}\cmidrule(lr){2-3}\cmidrule(lr){4-5}\cmidrule(lr){6-7}\cmidrule(lr){8-9}\cmidrule(lr){10-11}\cmidrule(lr){12-13}

TimesFM$_{\text{zero}}$ & 0.0272 & 0.0191 & 0.0447 & 0.0309 & 0.0669 & 0.0471 & 0.0264 & 0.0188 & 0.0432 & 0.0301 & 0.0660 & 0.0464 \\

TimesFM & 0.0244 & 0.0175 & 0.0392 & 0.0283 & 0.0582 & 0.0437 & 0.0237 & 0.0168 & 0.0380 & 0.0268 & 0.0569 & 0.0416 \\

Chronos2$_{\text{zero}}$ & 0.0166 & 0.0114 & 0.0328 & 0.0226 & 0.0543 & 0.0388 & 0.0177 & 0.0121 & 0.0357 & 0.0242 & 0.0696 & 0.0485 \\

Chronos2 & 0.0166 & 0.0114 & 0.0318 & 0.0220 & 0.0537 & 0.0392 & 0.0169 & 0.0115 & 0.0334 & 0.0223 & 0.0616 & 0.0432 \\
\bottomrule
\end{tabular}%
}
\caption{Performance comparison of models on multivariate house-price forecasting.}
\label{tab:performance_metrics_transposed}
\end{table*}

\begin{table*}[t]
\centering
\resizebox{\textwidth}{!}{%
\begin{tabular}{lcccccccccccc}
\toprule
\textbf{Window size $\rightarrow$} & \multicolumn{2}{c}{\{6,3\}} & \multicolumn{2}{c}{\{6,6\}} & \multicolumn{2}{c}{\{6,12\}} & \multicolumn{2}{c}{\{12,3\}} & \multicolumn{2}{c}{\{12,6\}} & \multicolumn{2}{c}{\{12,12\}} \\
\cmidrule(lr){2-3}\cmidrule(lr){4-5}\cmidrule(lr){6-7}\cmidrule(lr){8-9}\cmidrule(lr){10-11}\cmidrule(lr){12-13}
\textbf{Model $\downarrow$} & Log\textendash RMSE & MAPE & Log\textendash RMSE & MAPE & Log\textendash RMSE & MAPE & Log\textendash RMSE & MAPE &  Log\textendash RMSE & MAPE & Log\textendash RMSE & MAPE \\
\midrule

AR & 0.0256 & 0.0177 & 0.0432 & 0.0296 & 0.0891 & 0.1957 & 0.0256 & 0.0177 & 0.0433 & 0.0296 & 0.0731 & 0.0499 \\

RandomForest & 0.0182 & 0.0120 & 0.0344 & 0.0234 & 0.0593 & 0.0438 & 0.0191 & 0.0127 & 0.0350 & 0.0247 & 0.0589 & 0.0439 \\

XGBoost & 0.0373 & 0.0192 & 0.0464 & 0.0230 & 0.0640 & 0.0456 & 0.0362 & 0.0187 & 0.0445 & 0.0283 & 0.0653 & 0.0452 \\

\cmidrule(lr){1-1}\cmidrule(lr){2-3}\cmidrule(lr){4-5}\cmidrule(lr){6-7}\cmidrule(lr){8-9}\cmidrule(lr){10-11}\cmidrule(lr){12-13}

RNN & 0.0694 & 0.0572 & 0.0326 & 0.0233 & \colorbox{pink!30}{0.0537} & \colorbox{pink!30}{0.0395} & 0.0352 & 0.0262 & 0.0442 & 0.0360 & 0.1151 & 0.0846 \\

LSTM & 0.0140 & 0.0103 & \colorbox{pink!30}{0.0293} & \colorbox{pink!30}{0.0206} & 0.0606 & 0.0460 & 0.0135 & 0.0098 & \colorbox{pink!30}{0.0300} & \colorbox{pink!30}{0.0202} & 0.0596 & 0.0444 \\

STGCN & 0.0525 & 0.0432 & 0.0620 & 0.0501 & 0.0797 & 0.0606 & 0.0516 & 0.0438 & 0.0535 & 0.0409 & 0.0913 & 0.0751 \\

Graph WaveNet & 0.0308 & 0.0239 & 0.0502 & 0.0416 & 0.0646 & 0.0497 & 0.0292 & 0.0224 & 0.0467 & 0.0356 & 0.0729 & 0.0579 \\

DLinear & \colorbox{red!30}{0.0110} & \colorbox{red!30}{0.0071} & \colorbox{red!30}{0.0263} & \colorbox{red!30}{0.0173} & \colorbox{red!30}{0.0517} & \colorbox{red!30}{0.0377} & \colorbox{red!30}{0.0112} & \colorbox{red!30}{0.0069} & \colorbox{red!30}{0.0270} & \colorbox{red!30}{0.0175} & \colorbox{red!30}{0.0531} & \colorbox{red!30}{0.0378} \\

Autoformer & 0.0234 & 0.0175 & 0.0376 & 0.0282 & 0.0790 & 0.0700 & 0.0351 & 0.0308 & 0.0384 & 0.0291 & 0.0697 & 0.0597 \\

PatchTST & 0.0247 & 0.0199 & 0.0385 & 0.0296 & 0.0692 & 0.0541 & 0.0407 & 0.0341 & 0.0485 & 0.0391 & 0.0679 & 0.0519 \\

FEDformer & 0.0237 & 0.0179 & 0.0371 & 0.0274 & 0.0585 & 0.0449 & 0.0252 & 0.0199 & 0.0469 & 0.0400 & 0.0588 & 0.0449 \\

Informer & 0.0339 & 0.0260 & 0.0513 & 0.0390 & 0.0662 & 0.0533 & 0.0402 & 0.0343 & 0.0445 & 0.0344 & 0.0728 & 0.0600 \\

TimeMixer & \colorbox{pink!30}{0.0126} & \colorbox{pink!30}{0.0087} & 0.0303 & 0.0223 & 0.0635 & 0.0475 & \colorbox{pink!30}{0.0113} & \colorbox{pink!30}{0.0077} & 0.0312 & 0.0232 & \colorbox{pink!30}{0.0573} & 0.0426 \\

\cmidrule(lr){1-1}\cmidrule(lr){2-3}\cmidrule(lr){4-5}\cmidrule(lr){6-7}\cmidrule(lr){8-9}\cmidrule(lr){10-11}\cmidrule(lr){12-13}

TimesFM$_{\text{zero}}$ & 0.0274 & 0.0190 & 0.0466 & 0.0319 & 0.0695 & 0.0488 & 0.0202 & 0.0139 & 0.0400 & 0.0269 & 0.0724 & 0.0496 \\

TimesFM & 0.0244 & 0.0174 & 0.0400 & 0.0287 & 0.0594 & 0.0443 & 0.0182 & 0.0124 & 0.0352 & 0.0238 & 0.0602 & \colorbox{pink!30}{0.0418} \\

Chronos2$_{\text{zero}}$ & 0.0163 & 0.0111 & 0.0328 & 0.0226 & 0.0558 & 0.0401 & 0.0175 & 0.0120 & 0.0346 & 0.0236 & 0.0684 & 0.0477 \\

Chronos2 & 0.0163 & 0.0112 & 0.0324 & 0.0223 & 0.0571 & 0.0413 & 0.0170 & 0.0115 & 0.0335 & 0.0225 & 0.0675 & 0.0469 \\
\bottomrule
\end{tabular}%
}
\caption{Performance comparison of models on Univariate house-price data forecasting.}
\label{tab:univariate_price_only}
\end{table*}

}

\textbf{Points of interest.}
POI capture the presence of local amenities and services within each ZIP code and provide a proxy for neighborhood function and accessibility. We compile monthly amenity counts from March 2012 to December 2023 using historical OpenStreetMap data accessed through the OSHDB framework~\cite{raifer2019oshdb}. The POI categories include banks, bus-related facilities, hospitals, shopping malls, parks, restaurants, schools, transit stations, and supermarkets. For each ZIP code, we determine the corresponding geographic area and aggregate POIs within that area at monthly resolution.

\textbf{Census data.}
HouseTS includes socioeconomic variables from the American Community Survey obtained through the U.S.\ Census Bureau API~\cite{ACS2025}. The data span 2011 through 2022 and provide annual ZIP-code-level estimates of demographic and economic conditions. Included variables cover population and age structure, income and poverty measures, housing stock and housing costs, labor-force participation and unemployment, school-age population and enrollment, and commuting time. As described in the alignment protocol, ACS features are forward-shifted when paired with monthly targets to ensure temporal validity and prevent leakage. We provide feature target correlation summaries for POI and census variables in Appendix Figure~\ref{fig:5}.

\textbf{Aerial imagery and textual annotations.}
HouseTS includes aerial imagery from the National Agriculture Imagery Program~\cite{usda_naip}. We provide one RGB image per ZIP code per year from 2012 to 2022. These images are indexed by ZIP code and year and are released as an auxiliary modality for multimodal analysis and interpretability studies. Appendix Figure~\ref{fig:6} shows an example image and illustrates the diversity of geographic contexts covered by the dataset. Because collection frequency and coverage vary across states and years, not every ZIP code has imagery for every year. We preserve the imagery as released and do not impute missing years.
We also provide image-derived textual change annotations based on the time-stamped imagery. Vision--language models convert multi-year image sequences into structured neighborhood-change descriptors. The generated descriptors are quality-controlled using an LLM as a judge and then verified by human annotators, as described in Section~\ref{sec:vlm_semantic_layer}. These annotations support qualitative inspection of neighborhood evolution and enable multimodal analyses alongside the tabular benchmark.

\textbf{Data preprocessing.}
HouseTS uses a consistent preprocessing pipeline for all tabular modalities. All preprocessing statistics are fit on the training split only, and forward-filling is applied within each ZIP code after splitting. Missing values arise from incomplete coverage across sources and time. We handle missingness with a time-aware imputation strategy. Within each ZIP-code time series, we propagate observations forward in time to fill short gaps. Remaining missing entries are filled using ZIP-level medians computed on the training split, and any residual missingness is filled using global feature-wise medians computed on the training split. Negative placeholder values that encode upstream missingness are set to zero before imputation, while genuine zeros such as the absence of a POI category are preserved. To reduce skew and stabilize learning, we apply a logarithmic transform with a unit offset to continuous variables in the housing-market, POI, and census modalities. Figure~\ref{price} illustrates the heavy right-skew of raw prices and the substantially more symmetric distribution after transformation. Additional descriptive statistics for all covariates are reported in Appendix Table~\ref{tab:desc_stats}. We train and evaluate forecasting models in the transformed space and apply the inverse transform only when reporting results in original units.

\textbf{Additional tasks enabled by HouseTS.} Beyond the primary forecasting setting, HouseTS also enables multi-target prediction, structured-missingness, cross-city transfer with held-out metropolitan areas, and multimodal change understanding using time-stamped aerial imagery. We also release the raw, unimputed source tables to support alternative preprocessing choices. We leave a systematic study of these extensions to future work.

\textbf{Ethics and fairness.}
HouseTS is constructed from publicly available sources and contains no personally identifiable information. All tabular variables are ZIP-code-level aggregates, and the imagery consists of publicly released aerial photographs. The dataset does not include direct human-subject data collection, and we rely on the consent and usage terms provided by the original data publishers. We document privacy considerations, data provenance, licensing and attribution requirements, known limitations and potential biases, and responsible-use guidance in Appendix~\ref{app:ethics_license}.

\section{Benchmark}

We benchmark a broad set of forecasting models on HouseTS to establish strong and reproducible baselines. We consider 16 model families spanning (i) statistical and classical machine learning baselines, (ii) deep neural networks,  and (iii) pretrained time-series foundation models. We also report variants such as zero-shot and fine-tuned foundation models.


\subsection{Benchmark evaluation methodology}
\label{sec:baseline_method}

We implement and evaluate all baselines under a unified protocol to ensure comparability. Let $i$ index ZIP codes and let $t$ index months. For each ZIP code, we observe a multivariate time series $\mathbf{x}_{t,i}\in\mathbb{R}^{F}$ and a scalar target series $y_{t,i}\in\mathbb{R}$ that represents house price. Given an input window length $L$ and a forecast horizon $H$, the forecasting task is
\begin{equation}
\widehat{\mathbf{y}}_{t+1:t+H,i}
= f_{\theta}\!\left(\mathbf{x}_{t-L+1:t,i}\right),
\label{eq:task_def}
\end{equation}
where $\mathbf{x}_{t-L+1:t,i}=\{\mathbf{x}_{t-L+1,i},\dots,\mathbf{x}_{t,i}\}$ and $\widehat{\mathbf{y}}_{t+1:t+H,i}\in\mathbb{R}^{H}$ denotes the $H$-step-ahead predictions for ZIP code $i$. We evaluate two input settings. In the univariate setting, the input is the target history so that $\mathbf{x}_{t,i}=y_{t,i}$. In the multivariate-input setting, $\mathbf{x}_{t,i}$ contains all continuous covariates while the target remains $y_{t,i}$. We use six window configurations with input length $L$ chosen from 6 and 12 months and horizon $H$ chosen from 3, 6, and 12 months. All models are trained and evaluated in log space using $\tilde{y}_{t,i}=\log(1+y_{t,i})$ and optimized with mean squared error. We report Log-RMSE in log space and MAPE on the original scale after applying the inverse transform. All methods use the same preprocessing pipeline, data splits, and window settings. We perform minimal cleaning and core model architectures are not modified.



\textbf{Statistical and traditional machine learning models.}
We include lightweight statistical and classical machine learning baselines that operate directly on the preprocessed tabular series. For the \textit{univariate} task, we use an autoregressive model fitted on the log-transformed target history to produce multi-step forecasts. For the \textit{multivariate-to-single} task, we use an autoregressive distributed lag baseline, implemented as direct multi-step ridge regression on lagged multivariate inputs, which preserves the task semantics and does not require future covariates. In addition, we evaluate two tree-based regressors, Random Forest and XGBoost, trained on lagged features using a direct multi-output strategy that maps a fixed-length input window to the full forecast horizon. 

\textbf{Deep learning and foundation models.}
Our deep learning baselines include both sequence models and spatiotemporal graph neural networks. For graph-based models, we evaluate STGCN and Graph WaveNet, which capture cross-ZIP interactions through a graph structure, with graph construction details reported in Appendix~\ref{app:gnn}. We also evaluate pretrained time-series foundation models in both zero-shot inference and task-adapted fine-tuning. Because these models are designed for univariate forecasting, we extend them to multivariate targets by applying the model separately to each target channel and stacking the resulting forecasts to match the desired output dimension. To ensure fair comparison, we apply a fixed and comparable hyperparameter search across baselines and select models based on validation performance.



\begin{table}[t]
\centering
\setlength{\tabcolsep}{5pt}
\renewcommand{\arraystretch}{1.15}

\begin{tabular}{lcccc}
\toprule
\textbf{Model $\downarrow$} &
\multicolumn{2}{c}{\textbf{log}} &
\multicolumn{2}{c}{\textbf{log + z-score}} \\
\cmidrule(lr){2-3}\cmidrule(lr){4-5}
& \textbf{Log--RMSE} & \textbf{MAPE} & \textbf{Log--RMSE} & \textbf{MAPE} \\
\midrule
ARDL          & 0.0302 & 0.0212 & 0.0297 & 0.0207 \\
RandomForest  & 0.0383 & 0.0278 & 0.0383 & 0.0278 \\
XGBoost       & 0.0431 & 0.0287 & 0.0429 & 0.0289 \\
\cmidrule(lr){1-5}
RNN           & 0.0731 & 0.0589 & 0.0431 & 0.0314 \\
LSTM          & 0.0624 & 0.0506 & 0.0356 & 0.0264 \\
DLinear       & 0.0399 & 0.0312 & 0.0291 & 0.0200 \\
Informer      & 0.6481 & 0.4567 & 0.0554 & 0.0447 \\
TimeMixer     & 0.1605 & 0.1411 & 0.0360 & 0.0276 \\
PatchTST      & 0.3041 & 0.2447 & 0.0486 & 0.0378 \\
Autoformer    & 0.0489 & 0.0412 & 0.0418 & 0.0324 \\
FEDformer     & 0.0426 & 0.0336 & 0.0442 & 0.0352 \\
STGCN         & 0.2578 & 0.2395 & 0.0756 & 0.0627 \\
Graph WaveNet & 0.1753 & 0.1352 & 0.0486 & 0.0385 \\
\cmidrule(lr){1-5}
TimesFM$_{\text{zero}}$         & 0.0460 & 0.0322 & 0.0458 & 0.0321 \\
TimesFM                         & 0.0404 & 0.0293 & 0.0401 & 0.0292 \\
Chronos2$_{\text{zero}}$         & 0.0427 & 0.0319 & 0.0378 & 0.0263 \\
Chronos2                         & 0.0408 & 0.0308 & 0.0357 & 0.0249 \\
\bottomrule
\end{tabular}

\caption{Multivariate preprocessing ablation, each entry is the mean over all forecasting horizons.}
\label{tab:mv_preproc_log_vs_zscore}
\end{table}

\subsection{Benchmark evaluation results}
Table~\ref{tab:performance_metrics_transposed} summarizes multivariate forecasting performance, and Table~\ref{tab:univariate_price_only} reports the univariate setting where only the price history is available. Several consistent patterns emerge across both Log-RMSE and MAPE and across all window configurations. First, simple linear structure dominates this benchmark. DLinear achieves the lowest error in nearly all multivariate columns and is also the strongest performer in the univariate table. ARDL is a consistently competitive multivariate baseline and is closest to DLinear overall. It is the best or tied best classical baseline and remains strong on long horizons. Together, these results suggest that after appropriate preprocessing, much of the signal in ZIP-level house-price dynamics can be captured by low-complexity, strongly regularized, and largely linear temporal mappings. Increasing model capacity does not necessarily translate into better generalization for this dataset. Consistent with this trend, the foundation models are competitive but do not surpass the strongest lightweight baselines.

Second, multivariate covariates help classical baselines, but nonlinear tree ensembles are not sufficient to close the gap. In the multivariate setting, ARDL improves over tree-based regressors at short and mid horizons, while RandomForest and XGBoost remain solid but are consistently behind the best performers as the horizon increases. In the univariate setting, the AR baseline degrades with horizon length and shows large percentage errors at longer horizons, highlighting the limitations of autoregressive extrapolation when uncertainty compounds over multi-step prediction.

Third, preprocessing is a primary determinant of stability for neural baselines. The ablation in Table~\ref{tab:mv_preproc_log_vs_zscore} shows that adding z-score normalization on top of the log transform has only a marginal effect on tree-based models, but can be decisive for gradient-trained architectures. Several neural models are unstable under log-only preprocessing and become well-behaved once z-scoring is applied, such as Informer, TimeMixer, and PatchTST. This indicates that in multivariate house-price forecasting, careful normalization is a key methodological choice that can dominate model comparisons, aligning with the mixed-scale and heavy-tailed covariates in HouseTS that can otherwise yield ill-conditioned optimization. Finally, the graph-based approaches underperform relative to non-graph baselines across most horizons, indicating limited benefit from our simple fixed geographic kNN graph specification; richer or learned spatial priors may be necessary to better capture cross-ZIP interactions. Foundation models are competitive and generally robust, and fine-tuning improves over their zero-shot variants in several settings, but they do not surpass the strongest non-foundation baselines in this benchmark.

\section{Image-Derived Textual Annotations}
\label{sec:vlm_semantic_layer}

To support interpretability and multimodal analysis, we provide image-derived textual annotations for the time-stamped aerial imagery in HouseTS. Vision-language models convert multi-year image sequences into structured neighborhood descriptors that include both a long-term change summary and a description of the most recent visible state. The resulting annotations are quality-controlled using an LLM as a judge, and selected cases are verified by human annotators. We also release reliability metadata derived from cross-model disagreement to enable reliability-aware dataset usage.

\begin{figure}[t]
  \centering
  \includegraphics[width=\linewidth]{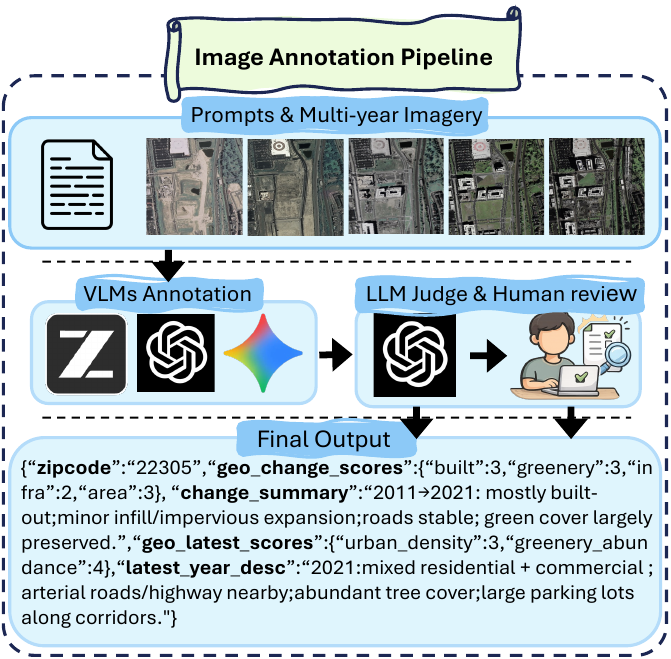}
  \caption{VLM-based semantic annotation pipeline.}
  \label{fig:vlm_pipeline}
\end{figure}

\subsection{VLMs aerial image annotations}
For each ZIP code, the VLM annotators receive the available annual aerial images and first sort them chronologically before reasoning about multi-year change. Each annotator then produces one structured record under a fixed schema. The record includes a concise summary that compares the earliest year with the latest usable year, a set of discrete scores that quantify change in the built environment, vegetation, and infrastructure, and an approximate changed-area ratio. It also includes a description of the most recent visible state together with current-state scores that characterize urban density and greenery abundance, plus an overall uncertainty score. The prompt enforces visual grounding by requiring that claims rely only on visible evidence, that uncertainty is stated when imagery is ambiguous, and that socioeconomic or demographic inferences are avoided. Figure~\ref{fig:vlm_pipeline} summarizes the overall annotation pipeline and the released outputs.

Multi-year aerial imagery may include unusable frames due to occlusion, blur, partial coverage, or corruption. To ensure that current-state annotations rely on valid evidence, we use a reference-year mechanism. The nominal latest year is the most recent year available for a ZIP code, while current-state descriptions and scores come from the most recent usable image. If the nominal latest image is unusable, the annotator falls back to the nearest earlier usable year and records a short rejection reason. If a ZIP code has only one usable image, we omit change summaries and change scores and report only the current-state description, current-state scores, and uncertainty to avoid unsupported change narratives.

\begin{figure}[t]
  \centering
  \includegraphics[width=\linewidth]{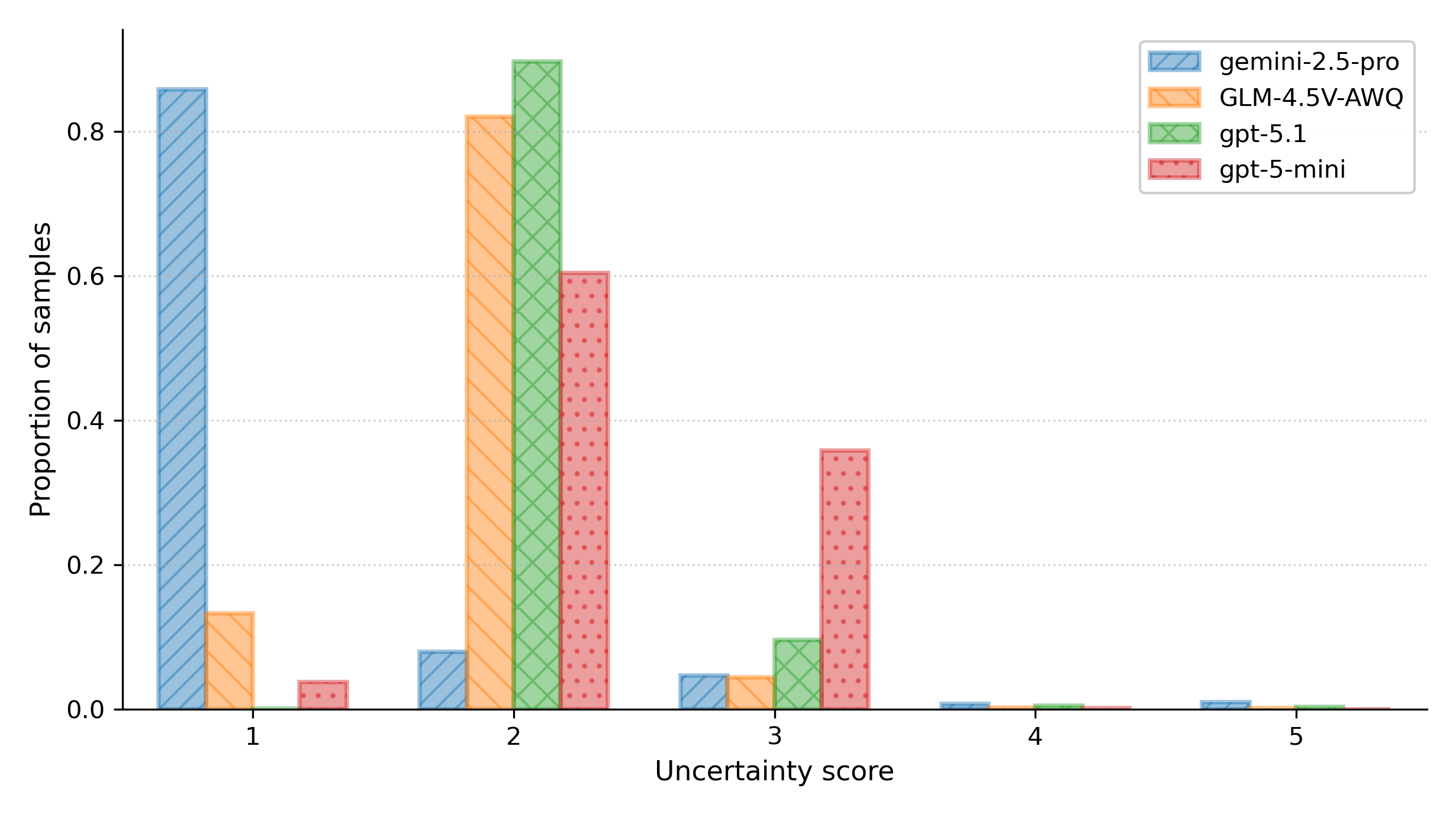}
  \caption{Model-specific distributions of self-reported uncertainty scores (1--5).}
  \label{fig:uncertainty_by_model}
\end{figure}

VLM outputs can vary across models. We therefore use multiple VLMs as independent annotators under the same prompt and schema, and we retain the per-model outputs for each ZIP code. Figure~\ref{fig:uncertainty_by_model} shows clear calibration differences in self-reported uncertainty. Gemini-2.5-pro assigns uncertainty score 1 for the majority of samples, while GLM-4.5V-AWQ and GPT-5.1 concentrate most samples at score 2. GPT-5-mini assigns higher uncertainty more often, with substantial mass at score 3. These shifts indicate that uncertainty scores are not directly comparable across annotators, which motivates model-agnostic reliability metadata based on cross-model disagreement.

\subsection{Disagreement-based reliability with LLM-as-a-judge and human verification}

\begin{figure}[t]
  \centering
  \includegraphics[width=\linewidth]{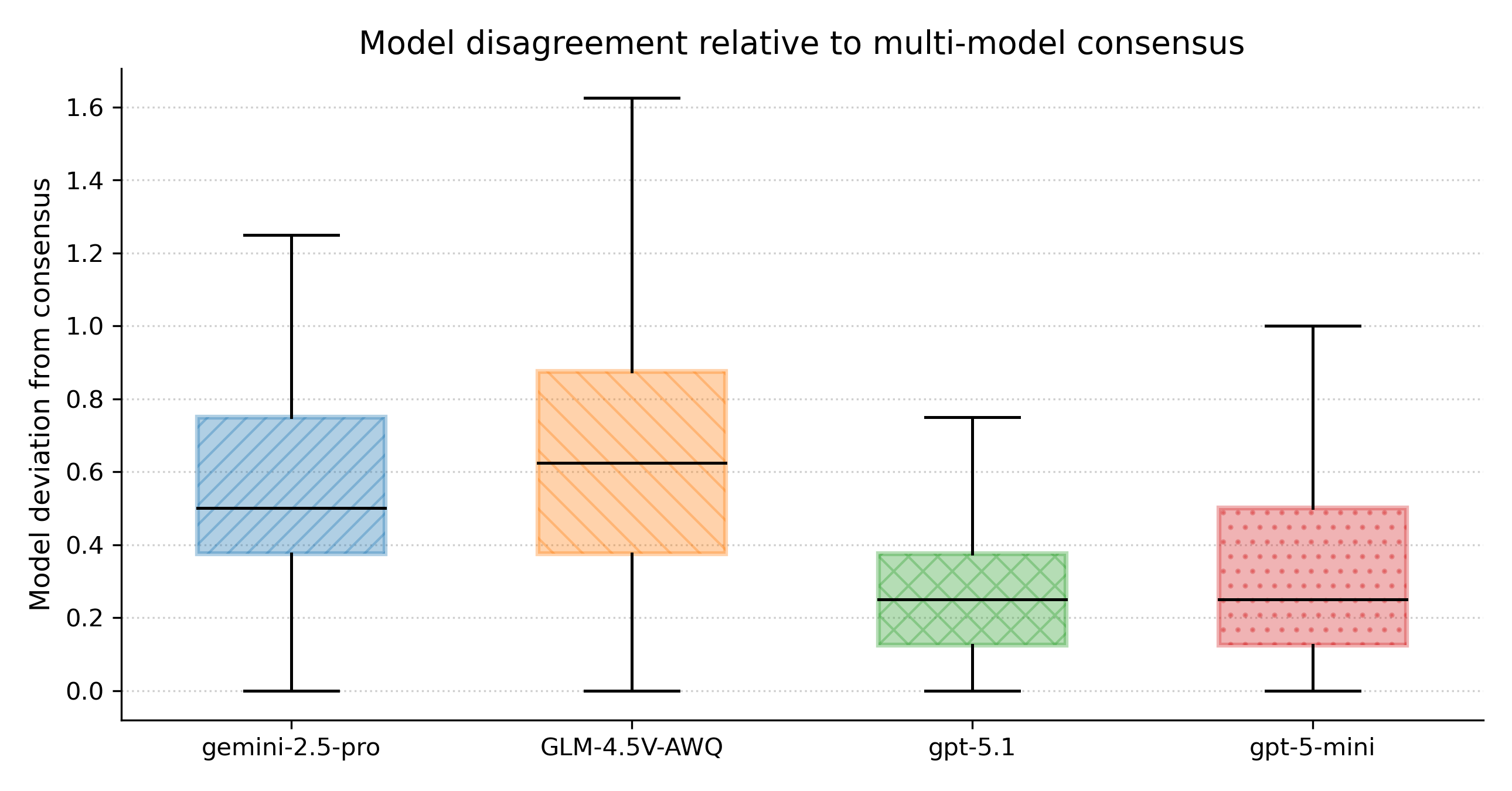}
  \caption{Deviation of each annotator relative to the multi-model consensus (median), measured as mean absolute difference over discrete score fields. Lower deviation indicates closer alignment to consensus.}
  \label{fig:model_deviation}
\end{figure}

To support downstream use, we aggregate discrete score fields across annotators using the median and release the resulting consensus labels. We also provide disagreement-based reliability metadata that summarizes cross-model variation for each sample, including dispersion and the maximum score gap across annotators. Figure~\ref{fig:model_deviation} summarizes annotator-specific deviation from the consensus, measured as the mean absolute difference over discrete score fields. The plot shows that models differ in how closely their scores track the median consensus, which motivates reporting both the consensus labels and per-model deviation statistics rather than a single merged output. 

\begin{figure}[t]
  \centering
  \includegraphics[width=\linewidth]{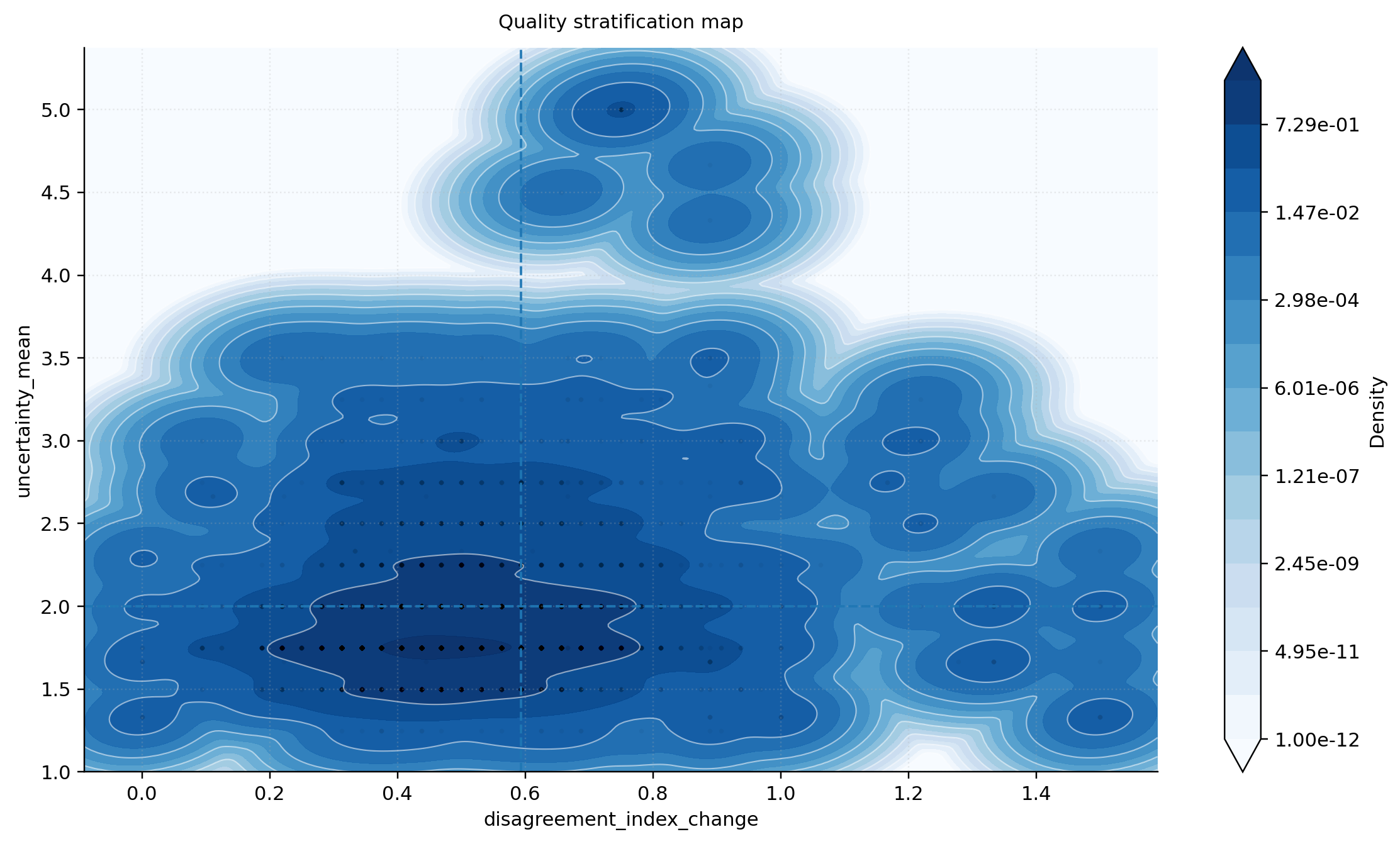}
  \caption{Quality stratification map based on disagreement and uncertainty. Contour density highlights a dominant high-confidence mass and a long tail of ambiguous samples.}
  \label{fig:quality_stratification}
\end{figure}

We treat cross-model disagreement as a dataset-level reliability signal and use it to stratify samples into confidence tiers for downstream use. Figure~\ref{fig:quality_stratification} plots the joint distribution of a disagreement index for change annotations and the mean self-reported uncertainty across annotators. Most samples concentrate in a dense region with relatively low disagreement, while a sparser tail extends toward higher disagreement and higher uncertainty, indicating cases where annotators diverge and visual evidence is more ambiguous. This stratification supports targeted quality control and budgeted human verification, since manual review can focus on the high-disagreement region rather than auditing the full dataset. We release both disagreement-based metadata and model-reported uncertainty scores to support reliability-aware filtering and analysis, even when uncertainty calibration differs across annotators.

To improve annotation reliability, we use an LLM-as-a-judge to decide which cases require human review. The judge model is GPT-5.2. It inspects the per-model annotation records and the reliability metadata. The judge prioritizes cases with high cross-model disagreement on discrete score fields and cases where annotator texts repeatedly indicate weak visual evidence, such as missing coverage, corrupted frames, or unclear imagery. Based on these signals, GPT-5.2 flags a subset for human verification and provides a brief rationale for each flag. For each flagged ZIP code, a reviewer inspects the multi-year images and compares the outputs from the four VLM annotators. The reviewer checks reference-year validity, the directional correctness of the change summary, and the reasonableness of the discrete scores, and then selects the most credible annotator output as the final annotation for that case. Unflagged samples keep the automatically produced consensus labels.

We release the image-derived textual annotations as part of the HouseTS dataset artifacts. For each ZIP code, we provide the multi-year aerial images, the per-model annotation records, and an aggregated file that contains median-based consensus labels together with disagreement-based reliability metadata. For cases selected for human verification, we provide the human-selected annotator output in the same schema and the LLM-as-a-judge rationale as accompanying metadata, enabling direct comparison with per-model outputs and consensus labels. The imagery and image-derived annotations complement the tabular benchmark. The main forecasting benchmark in this paper is defined on the aligned tabular modalities. The imagery and image-derived annotations complement this benchmark by supporting interpretability, multimodal extensions, and reliability-aware subset construction. We recommend using the reliability metadata to filter a high-confidence subset, to construct challenging subsets from high-disagreement samples, and to incorporate the descriptors and reliability signals as auxiliary covariates for multimodal analysis.

\section{Conclusion}
We present HouseTS, a ZIP-code-level multimodal spatiotemporal dataset for long-horizon housing-market analysis. HouseTS provides a unified monthly tabular panel for 6{,}000 ZIP codes across 30 major U.S.\ metropolitan areas from 2012 to 2023, integrating housing-market signals, neighborhood amenity dynamics, and annual socioeconomic indicators under a consistent alignment and preprocessing protocol, and it also includes time-stamped annual aerial imagery as an auxiliary modality for interpretability and multimodal research. Using HouseTS, we establish a standardized long-horizon forecasting benchmark in univariate and multivariate settings and report strong baselines spanning naive persistence, statistical methods, classical machine learning models, deep neural networks, and pretrained time-series foundation models. We further release image-derived textual annotations that describe both long-term neighborhood change and the most recent visible state, together with disagreement-based reliability metadata to support reliability-aware subset construction and multimodal extensions. We release the dataset, benchmarking code, and documentation to facilitate reproducible evaluation and future research on forecasting and related spatiotemporal learning problems that benefit from aligned market, socioeconomic, and amenity signals.


\renewcommand{\refname}{\MakeUppercase{References}}
\bibliographystyle{ACM-Reference-Format}
\bibliography{custom}

\clearpage
\appendix

\section{Feature Analysis}\label{appendix:feature-analysis}
\FloatBarrier

\begin{center}
\footnotesize
\resizebox{\columnwidth}{!}{%
\begin{tabular}{@{}lrrrrrrr@{}}
  \toprule
  \textbf{Feature} & \textbf{Mean} & \textbf{Std} & \textbf{Min} & \textbf{25\%} & \textbf{50\%} & \textbf{75\%} & \textbf{Max} \\ \midrule
  Median Age                       & 36.734 & 12.437 & 0.000 & 34.100 & 38.500 & 43.100 & 91.200 \\
  Median Commute Time              & 9,\,687.920 & 8,\,841.743 & 0.000 & 1,\,945.000 & 7,\,830.500 & 15,\,013.000 & 60,\,956.000 \\
  Median Home Value                & 314,\,792.244 & 267,\,219.362 & 0.000 & 143,\,600.000 & 250,\,100.000 & 412,\,500.000 & 2,\,000,\,001.000 \\
  Median Rent                      & 1,\,146.686 & 547.237 & 0.000 & 852.000 & 1,\,114.000 & 1,\,446.000 & 3,\,501.000 \\
  Per Capita Income                & 35,\,253.501 & 21,\,555.152 & 0.000 & 23,\,210.000 & 32,\,025.000 & 44,\,198.000 & 465,\,868.000 \\
  Total Families Below Poverty     & 21,\,457.525 & 19,\,554.170 & 0.000 & 4,\,404.000 & 17,\,489.000 & 32,\,991.000 & 130,\,605.000 \\
  Total Housing Units              & 8,\,714.481 & 7,\,588.635 & 0.000 & 1,\,930.000 & 7,\,426.000 & 13,\,564.000 & 48,\,734.000 \\
  Total Labor Force                & 11,\,455.780 & 10,\,429.516 & 0.000 & 2,\,320.000 & 9,\,299.000 & 17,\,702.000 & 68,\,735.000 \\
  Total Population                 & 21,\,802.546 & 19,\,794.374 & 0.000 & 4,\,512.000 & 17,\,848.000 & 33,\,538.000 & 130,\,920.000 \\
  Total School Age Population      & 20,\,998.338 & 19,\,008.391 & 0.000 & 4,\,369.000 & 17,\,247.000 & 32,\,290.000 & 126,\,948.000 \\
  Total School Enrollment          & 20,\,998.338 & 19,\,008.391 & 0.000 & 4,\,369.000 & 17,\,247.000 & 32,\,290.000 & 126,\,948.000 \\
  Unemployed Population            & 829.769 & 954.754 & 0.000 & 127.000 & 538.000 & 1,\,192.000 & 9,\,735.000 \\
  avg\_sale\_to\_list              & 0.978 & 0.064 & 0.000 & 0.965 & 0.982 & 0.998 & 1.906 \\
  bank                             & 13.384 & 31.045 & 0.000 & 0.000 & 4.000 & 15.000 & 447.000 \\
  bus                              & 0.670 & 1.610 & 0.000 & 0.000 & 0.000 & 1.000 & 26.000 \\
  homes\_sold                      & 76.723 & 76.698 & 0.000 & 19.000 & 55.000 & 111.000 & 955.000 \\
  hospital                         & 3.506 & 7.368 & 0.000 & 0.000 & 1.000 & 4.000 & 96.000 \\
  inventory                        & 77.301 & 89.042 & 0.000 & 20.000 & 50.000 & 103.000 & 1,\,941.000 \\
  mall                             & 1.292 & 2.752 & 0.000 & 0.000 & 0.000 & 1.000 & 45.000 \\
  median\_dom                      & 61.290 & 82.220 & 0.000 & 26.000 & 45.000 & 74.000 & 7,\,777.000 \\
  median\_list\_ppsf               & 231.170 & 290.120 & 0.000 & 116.818 & 173.143 & 270.181 & 143,\,015.399 \\
  median\_list\_price              & 422,\,984.881 & 1,\,899,\,201.111 & 0.000 & 199,\,000.000 & 320,\,000.000 & 499,\,900.000 & 999,\,999,\,999.000 \\
  median\_ppsf                     & 223.068 & 696.724 & 0.000 & 110.640 & 166.094 & 260.626 & 366,\,700.000 \\
  median\_sale\_price              & 394,\,102.626 & 381,\,548.138 & 0.000 & 185,\,000.000 & 302,\,500.000 & 480,\,000.000 & 20,\,500,\,000.000 \\
  new\_listings                    & 92.910 & 92.696 & 0.000 & 24.000 & 67.000 & 133.000 & 1,\,112.000 \\
  off\_market\_in\_two\_weeks       & 0.306 & 0.239 & 0.000 & 0.083 & 0.295 & 0.476 & 1.000 \\
  park                             & 48.989 & 75.719 & 0.000 & 5.000 & 24.000 & 63.000 & 926.000 \\
  pending\_sales                   & 81.471 & 85.328 & 0.000 & 17.000 & 57.000 & 119.000 & 1,\,374.000 \\
  price                            & 391,\,328.910 & 344,\,538.332 & 10,\,464.318 & 189,\,706.296 & 305,\,018.960 & 479,\,711.108 & 8,\,463,\,115.592 \\
  restaurant                       & 64.993 & 199.437 & 0.000 & 2.000 & 13.000 & 50.000 & 3,\,409.000 \\
  school                           & 48.667 & 62.302 & 0.000 & 7.000 & 27.000 & 66.000 & 560.000 \\
  sold\_above\_list                & 0.264 & 0.202 & 0.000 & 0.120 & 0.224 & 0.375 & 1.000 \\
  station                          & 5.703 & 16.774 & 0.000 & 0.000 & 0.000 & 4.000 & 192.000 \\
  supermarket                      & 9.718 & 19.202 & 0.000 & 1.000 & 4.000 & 12.000 & 303.000 \\
  \bottomrule
\end{tabular}}
\captionof{table}{Descriptive statistics for features.}
\label{tab:desc_stats}
\end{center}

Table~\ref{tab:desc_stats} summarizes descriptive statistics for the ZIP-level covariates and the price target. Many variables show substantial dispersion across ZIP codes and span wide numeric ranges, including housing and socioeconomic measures such as median home value and total population. Count-based amenity and transaction variables are often sparse and heavy-tailed, which motivates monotone variance-stabilizing transforms before modeling. The price target also exhibits a wide dynamic range, indicating that models must handle both cross-sectional heterogeneity and temporal variation.

\begin{center}
  \includegraphics[width=\columnwidth]{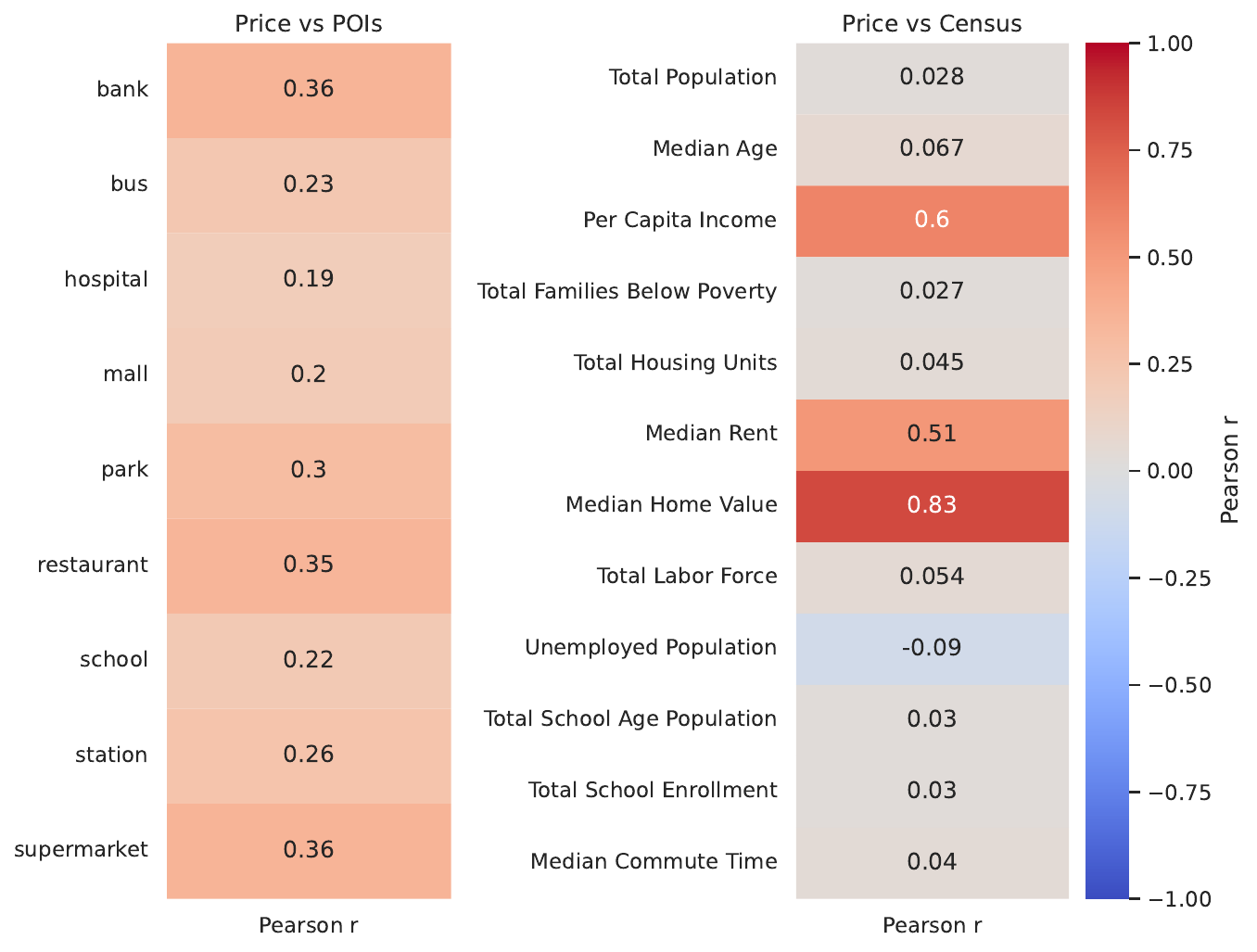}
  \captionof{figure}{Pearson $r$ correlations between median house price and POI densities and census variables.}
  \label{fig:5}
\end{center}

Figure~\ref{fig:5} reports Pearson correlations between the price target and representative covariates from the POI and census modalities. Several amenity-related POI counts show moderate positive correlations with house prices, including banks, restaurants, supermarkets, and parks, which is consistent with higher-priced ZIP codes tending to have denser services and amenities. Among census variables, median home value and median rent exhibit the strongest positive associations with price, and per capita income is also positively correlated. Most remaining demographic totals show comparatively weak linear correlations, and unemployed population is slightly negatively correlated.

\begin{center}
  \includegraphics[width=\columnwidth]{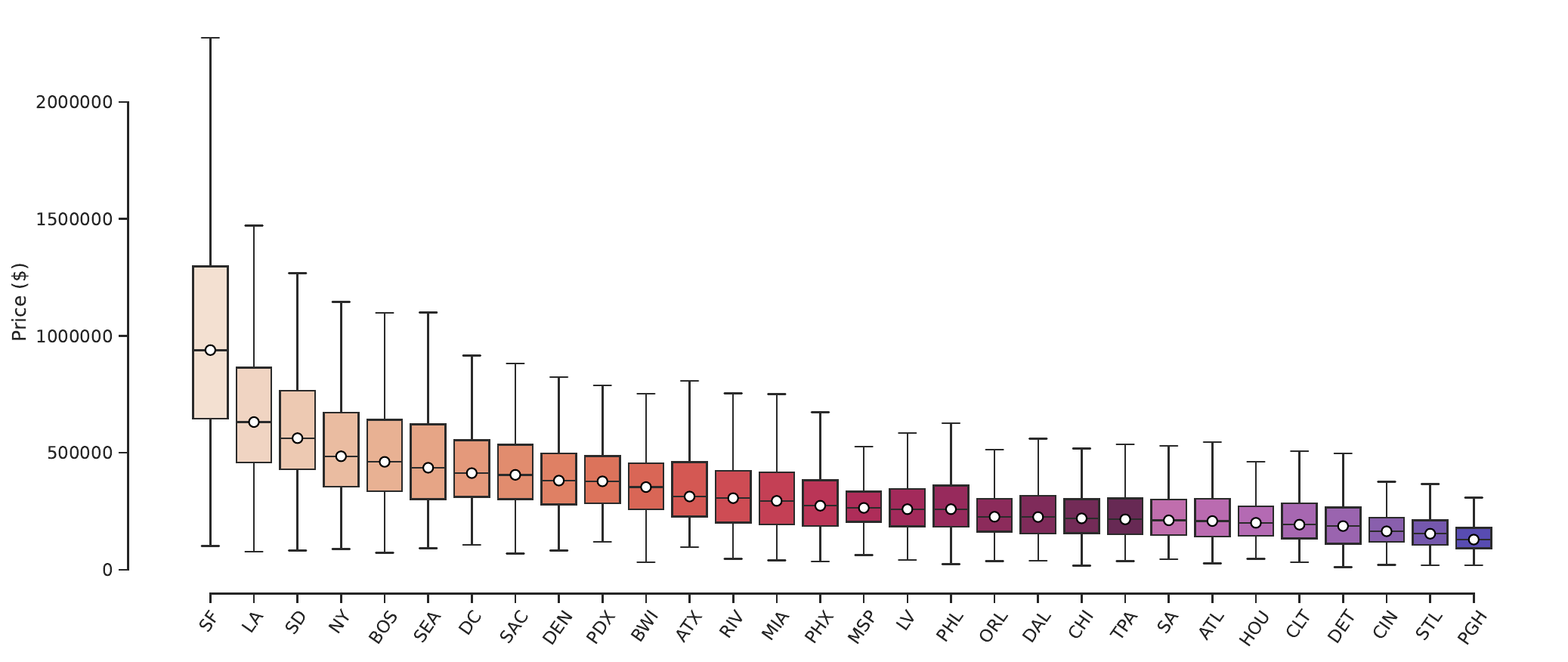}
  \captionof{figure}{House price distribution across regions.}
  \label{fig:4}
\end{center}

Figure~\ref{fig:4} shows boxplots of ZIP-level house prices grouped by metropolitan area. The distributions differ substantially across cities in both central tendency and dispersion. Some metropolitan areas have higher medians and wider interquartile ranges, while others concentrate at lower price levels with a tighter spread. Many cities also exhibit pronounced right tails and high-value outliers, indicating heavy-tailed price distributions within the same metropolitan area. This heterogeneity motivates models that can handle strong cross-sectional variation across locations and supports robust preprocessing such as log transforms to reduce sensitivity to extreme values.

\section{Exploratory PCA analysis}
\label{app:pca}
Although our reported forecasting baselines do not use PCA, we include this section as a descriptive reference to summarize the dominant variation patterns in the ZIP-level covariates. We fit PCA on the training split only, after applying the same preprocessing used for forecasting (e.g., log transform and normalization when enabled), and use the resulting projection for exploratory analysis and visualization.

We choose the number of retained components $k$ using a cumulative explained-variance criterion computed on training data. Specifically, we select the smallest $k$ such that
\[
\frac{\sum_{i=1}^{k}\lambda_i}{\sum_{j=1}^{D}\lambda_j} \ge 0.95,
\]
where $\lambda_i$ is the explained variance of component $i$ and $D$ is the original feature dimension. Under this criterion, we obtain $k=5$ on the training split. We emphasize that this choice is used only for consistent descriptive analysis (and any associated plots) and is not used in the forecasting baselines reported in the main paper.

Table~\ref{tab:pca_loadings} reports the absolute top loadings for the first three components, which provide a coarse interpretation of the covariate space. The first component is driven by broad socioeconomic and demographic magnitude signals . The second component is dominated by amenity and activity indicators. The third component reflects market dynamics variables, together with a value-related signal. Overall, the PCA projection reveals meaningful structure in the covariates and supports qualitative inspection of cross-sectional heterogeneity.

\begin{table}[t]
  \centering
  \begin{tabular}{@{}llc@{}}
    \toprule
    \textbf{PC} & \textbf{Feature} & \textbf{|Loading|} \\ \midrule
    \multirow{5}{*}{PC$_1$}
        & Median Home Value               & 0.350073 \\
        & Total Population                & 0.319604 \\
        & Total Families Below Poverty    & 0.319147 \\
        & Total School Age Population     & 0.318474 \\
        & Total School Enrollment         & 0.318474 \\ \midrule
    \multirow{5}{*}{PC$_2$}
        & restaurant                      & 0.382833 \\
        & park                            & 0.339089 \\
        & school                          & 0.315471 \\
        & bank                            & 0.271029 \\
        & pending\_sales                  & 0.259031 \\ \midrule
    \multirow{5}{*}{PC$_3$}
        & Median Home Value               & 0.329001 \\
        & restaurant                      & 0.323938 \\
        & inventory                       & 0.293042 \\
        & new\_listings                   & 0.291279 \\
        & homes\_sold                     & 0.286088 \\ \bottomrule
  \end{tabular}
  \caption{Absolute top-5 loadings of the first three principal components.}
  \label{tab:pca_loadings}
\end{table}

\vspace{-10pt}


\section{Geographic Graph Neural Network Baselines}
\label{app:gnn}

This appendix summarizes the graph construction and key hyperparameters for the spatiotemporal GNN baselines, which use a ZIP-code graph derived from public centroid coordinates. Let $V$ be the set of ZIP codes, with one node per ZIP. For each ZIP code $i$, we obtain its centroid coordinates and construct a geographic $k$-nearest-neighbor graph using Haversine distance. We include an edge from $i$ to $j$ if $j$ is among the $k$ nearest neighbors of $i$ and the distance is within an optional radius threshold:
\begin{equation}
E = \{(i,j) \mid j \in \mathrm{kNN}(i)\ \text{and}\ \mathrm{dist}(i,j) \le r\}.
\end{equation}
We make the graph undirected by adding reciprocal edges and include self-loops. Let $A$ denote the resulting sparse adjacency matrix. We apply symmetric normalization
\begin{equation}
\widehat{A} = D^{-\frac{1}{2}} A D^{-\frac{1}{2}},
\end{equation}
where $D$ is the degree matrix.

We use $k=10$ and $r=50$ km. Edges are unweighted and are normalized using the symmetric form above. For Graph WaveNet, we use the geographic graph as the fixed adjacency input and additionally enable its adaptive adjacency component, following the original formulation. We report the remaining model hyperparameters and training settings together with the full configuration files in the released codebase.

\section{Ethics, Fairness, and Data Licensing}
\label{app:ethics_license}

\noindent\textbf{Ethics and fairness.}
HouseTS is compiled from publicly available sources and contains no personally identifiable information: all tabular variables are aggregated at the ZIP-code level and the imagery consists of publicly released aerial photographs. The dataset does not recruit human participants and follows the access terms of the original publishers. As with other housing and demographic data, HouseTS may reflect historical inequities, representativeness gaps, measurement error, and sampling uncertainty, and coverage is limited to 30 major metropolitan areas with uneven source maintenance and imagery availability across states and years. We recommend reporting results both in aggregate and stratified by geography and socioeconomic conditions when feasible, and we discourage deployment in high-stakes decisions (e.g., housing access, credit, insurance, public services) without legal review, fairness auditing, and stakeholder oversight. We provide provenance and preprocessing documentation and encourage subgroup error analysis and clear statements of intended use when releasing trained artifacts.

\medskip
\noindent\textbf{Data license and source attributions.}
HouseTS is a compilation of multiple public sources, and each source retains its own terms. If any summary below conflicts with an upstream source policy, the upstream policy controls.

\emph{Zillow Research data.}
Zillow states that downloadable real estate metrics are free for public use subject to its Terms of Use and that clear attribution to Zillow is required.\footnote{\href{https://www.zillowgroup.com/developers/api/public-data/real-estate-metrics/}{Zillow Group Real Estate Metrics page}.}

\emph{Redfin Data Center.}
Redfin allows use of its downloadable housing market data and requests citation of the source, with a link to Redfin for the first reference on a page, post, or article.\footnote{\href{https://www.redfin.com/news/data-center/}{Redfin Data Center}.}

\emph{U.S. Census Bureau and the American Community Survey.}
ACS tabulations are published by the U.S.\ Census Bureau and are generally treated as public-domain government information. The Census Bureau requests appropriate citation.\footnote{\href{https://www.census.gov/about/policies/citation.html}{Census citation guidance}.}

\emph{OpenStreetMap history data and derived POI time series.}
POI statistics are derived from OpenStreetMap history data and require attribution to OpenStreetMap contributors.\footnote{\href{https://www.openstreetmap.org/copyright}{OpenStreetMap copyright and license}.}

\emph{USDA NAIP aerial imagery.}
NAIP imagery is U.S.\ government public domain data. Credit to USDA and the relevant distributor is requested when publishing derived imagery or products.\footnote{\href{https://www.usgs.gov/centers/eros/science/usgs-eros-archive-aerial-photography-national-agriculture-imagery-program-naip}{USGS NAIP overview page}.}
\footnote{\href{https://www.usgs.gov/information-policies-and-instructions/acknowledging-or-crediting-usgs}{USGS credit guidance}.}



\section{Aerial Image and Prompts for Image-Derived Textual Annotations}

\begin{center}
  \includegraphics[width=\columnwidth]{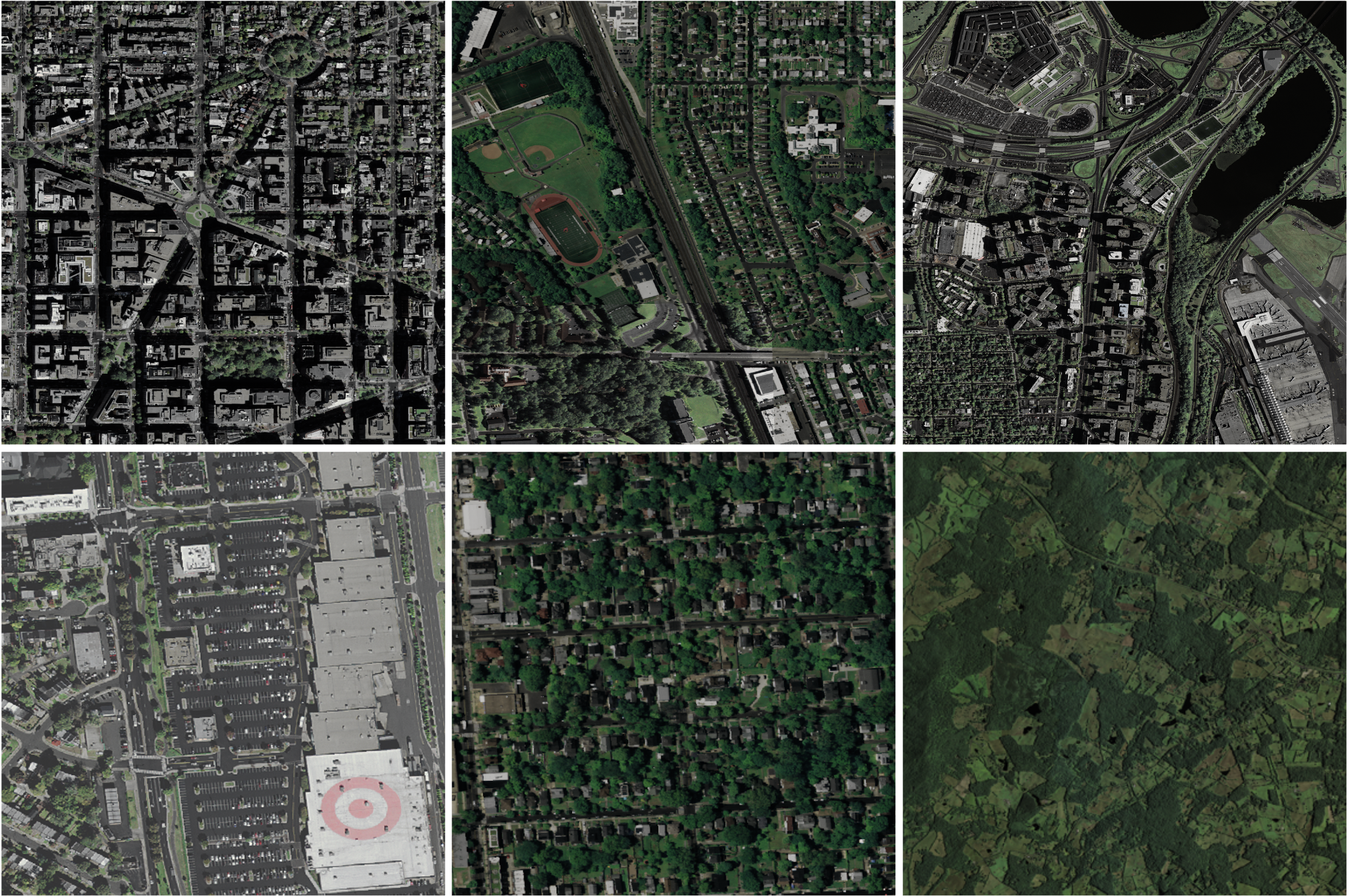}
  \captionof{figure}{Sample aerial image illustrating the dataset.}
  \label{fig:6}
\end{center}

We retrieve aerial imagery from NAIP via Google Earth Engine and use ZCTA boundaries to represent each 5-digit ZIP code. For each year and ZIP, we collect all NAIP frames that intersect the ZIP polygon, sort them by acquisition time in descending order, and build a single annual RGB. The composite is clipped to a buffered ZIP region ZCTA polygon with a 200\,m buffer and exported as a georeferenced \(512\times512\) GeoTIFF to provide a consistent input shape across ZIPs.

\begin{tcolorbox}[
  colback=gray!8,
  colframe=black,
  boxrule=0.4pt,
  left=2mm,right=2mm,top=1.5mm,bottom=1.5mm,
  breakable,
  enhanced,
  title={Prompt for image-derived textual annotations},
  fonttitle=\bfseries,
  listing only,
]
\footnotesize
You are an urban remote sensing expert.

You will receive N satellite images (.png) of the SAME ZIP code area in the United States.
Each image is a top-down aerial/satellite view.

The filenames of the images follow the pattern:
\textless zipcode\_year\textgreater.png

The year order in filenames is NOT guaranteed, so you MUST infer the years from the filenames and then sort the images from oldest year to newest year before you reason about changes.

Your tasks:

0. Special case: if N = 1 (single image only)
- Do NOT describe or score “change over time”.
- Set: first\_year = latest\_year = the single image year.
- Output only the current-state fields:
- latest\_year\_geography\_description
- latest\_urban\_density\_score, latest\_greenery\_abundance\_score
- uncertainty\_score

1. Understand the timeline
- Parse the year from each filename.
- Sort all images in chronological order from the earliest year to the latest year.
- Treat the first image in time as the "earliest" baseline and the last image as the "latest" state.
- You can assume all images belong to the same ZIP code and cover approximately the same area.
Select the reference year for “current state”:
- Let \texttt{latest\_year} be the maximum year that appears in the filenames (the most recent year available).
- You must choose a year to represent the “last state” for the geography description and latest-year scores.
- However, if the \texttt{latest\_year} image is NOT usable due to poor visibility or quality, you MUST choose the most recent earlier year that IS usable, i.e., the closest year to \texttt{latest\_year} with acceptable visibility.

2. Describe long-term visual changes (earliest $\rightarrow$ latest)
- Compare the earliest and latest images and summarize the main land-cover and built-environment changes over the whole time span.
- Focus only on what is clearly visible in the imagery, such as:
- Buildings / built-up area (residential, commercial, industrial structures)
- Roads and transportation infrastructure (highways, major roads, rail lines, large parking lots)
- Vegetation and greenery (trees, parks, fields, lawns)
- Large impervious surfaces (roofs, asphalt, concrete)
- Ignore fine-grained socio-economic attributes that cannot be directly seen.
- Write a concise paragraph (3–6 sentences) that explains how the area has changed from the earliest year to the latest year.

3. Assign four change-related quantitative scores (1–5)
Based ONLY on what you can clearly see in the images, assign FOUR INTEGER scores in the range 1–5 (inclusive):

- built\_environment\_change\_score (1–5):
- 1 = almost no visible change in buildings / built-up area
- 3 = moderate change (some new buildings or visible densification)
- 5 = very large change (many new buildings, major densification or redevelopment)

- greenery\_change\_score (1–5):
- 1 = large decrease in visible vegetation (trees/green areas clearly reduced)
- 3 = mixed or small net change
- 5 = large increase in visible vegetation (much more green cover than before)

- infrastructure\_change\_score (1–5):
- 1 = almost no visible change in roads or large engineered structures
- 3 = some new or widened roads / noticeable infrastructure additions
- 5 = major new infrastructure (new highways, interchanges, large facilities, etc.)

- changed\_area\_ratio\_score (1–5):
This reflects roughly how large a fraction of the visible area has undergone noticeable change between the earliest and latest images, regardless of whether the change is from green to built-up or built-up to green.
1 = very small portion of the area appears to have changed
3 = a moderate portion of the area appears to have changed
5 = a large portion of the area appears to have changed (widespread visible changes)

Use the full 1–5 range when appropriate. If you are uncertain for any reason
(e.g., large parts of the images are obscured, the resolution is low, the
evidence is ambiguous, or you cannot clearly judge the magnitude of change), you must:
- Explicitly mention this uncertainty in your explanation.
- Choose a conservative middle value (e.g., 2–4) instead of extreme values 1 or 5.

4. Describe the latest-year geography (current state description)
- Look ONLY at the image corresponding to the latest year.
- Write a short paragraph (3–6 sentences) describing the current geography and land use of this area, such as:
- Whether it is mainly residential, commercial, industrial, or mixed
- Presence of large roads or highways
- Amount and pattern of greenery (parks, tree cover, fields)
- Any notable visible features (large parking lots, big box stores, high-rise complexes, water bodies, etc.)
- Stay at a high level and use generic terms (e.g., “buildings”, “industrial facilities”, “large commercial buildings”, “parking lots”).
Do NOT guess specific brands, companies, building names, or detailed socio-economic conditions.
- This paragraph should describe only the latest-year image, NOT the changes over time.

5. Assign two latest-year scores (1–5)
Based ONLY on the LATEST-year image, assign TWO additional INTEGER scores in the range 1–5 (inclusive) that describe the current state:

- latest\_urban\_density\_score (1–5):
1 = very low density (mostly open land/vegetation with scattered buildings)
3 = moderate density (typical suburban or low-to-medium density mixed use)
5 = very high density (most of the area covered by buildings and impervious surfaces)

- latest\_greenery\_abundance\_score (1–5):
1 = very little greenery (mostly buildings/roads/impervious surfaces)
3 = moderate greenery (noticeable trees/parks but not dominant)
5 = abundant greenery (large share of the area covered by trees, parks, or other green areas)

6. Assign an overall uncertainty score (1–5)
Finally, assign ONE INTEGER uncertainty\_score (1–5) that reflects your overall uncertainty about your assessments for this ZIP code, considering image quality, coverage, and clarity of changes:

- uncertainty\_score (1–5):
1 = very low uncertainty (images are clear, changes are easy to interpret)
3 = moderate uncertainty (some parts are unclear, but you can still make reasonable judgments)
5 = very high uncertainty (large areas are unclear, or changes are extremely hard to interpret)

When uncertainty\_score is high, you should be more conservative with all change-related and latest-year scores and avoid extreme values (1 or 5) unless changes are clearly visible.

7. Grounding and hallucination constraints (very important)
- Use ONLY information that is clearly visible in the images and their filenames.
- Do NOT invent features that you cannot clearly see.
- If you are uncertain about something, clearly say that you are uncertain instead of guessing.
- Do NOT infer or mention:
- income, wealth, race, crime, safety, “gentrification”, or detailed demographics
- housing prices, rents, or other economic outcomes
- specific business names, brands, or institutions
- If image quality, clouds, shadows, or cropping prevent you from seeing parts of the area, explicitly state that some areas are not visible and base your scores only on the visible parts.

8.Required JSON output schema
- Your entire response MUST be a single valid JSON object.

- Do NOT include any extra commentary or text outside the JSON.

- Use this exact structure:

\{
"zipcode": "\textless the zipcode parsed from the filenames\textgreater",
"first\_year": YEAR\_FIRST,
"latest\_year": YEAR\_LAST,
"change\_summary": "\textless 3-6 sentence paragraph describing how the area changed from the earliest year to the latest year\textgreater",
"built\_environment\_change\_score": X,
"greenery\_change\_score": Y,
"infrastructure\_change\_score": Z,
"changed\_area\_ratio\_score": A,
"latest\_year\_geography\_description": "\textless 3-6 sentence paragraph describing the current geography/land use of the latest-year image only\textgreater",
"latest\_urban\_density\_score": B,
"latest\_greenery\_abundance\_score": C,
"uncertainty\_score": D
\}
\end{tcolorbox}

\begin{figure*}[t]
\centering
\begin{minipage}{0.98\textwidth}
\begin{tcolorbox}[
  colback=gray!8,
  colframe=black,
  boxrule=0.4pt,
  left=2mm,right=2mm,top=1.5mm,bottom=1.5mm,
  enhanced,
  fonttitle=\bfseries,
  title={LLM-as-a-judge prompt for annotation triage},
]
\footnotesize
\begin{verbatim}
You are a strict quality-control judge for a dataset annotation pipeline.

You will be given multiple VLM annotator outputs for the SAME US ZIP code area.
Each annotator output is a JSON object produced from satellite images across years.
You do NOT see the images. You ONLY see the annotators' JSON outputs.

Your goals:
1) Decide whether the annotations are reliable enough to be accepted WITHOUT human review.
2) If acceptable, produce ONE final annotation JSON that can be used as the dataset label.
3) If not acceptable, mark needs_human_review = true and explain the reason briefly.

You MUST apply these rules:

A. Agreement & consistency checks (cross-model)
- Compare first_year, latest_year, and all score fields.
- Check if change_summary and latest_year_geography_description are semantically consistent across models.
- If models describe clearly contradictory trends (e.g., one says "major development" another says "almost no change"),
  treat as disagreement.
- If models disagree strongly on infrastructure change (e.g., score gaps >= 2) AND their text rationales conflict,
  treat as disagreement.

B. Uncertainty-aware logic
- If most annotators report high uncertainty_score (>=4), or their texts repeatedly mention "cannot see / unclear",
  prefer needs_human_review = true unless consensus is still strong.
- If uncertainty is low (mostly 1–2), you can be more confident in accepting.

C. Final annotation construction (when accepted)
- Use a consensus strategy:
  - For each numeric score: use the median across available models (round to nearest integer, clamp 1..5).
  - For first_year/latest_year: use the median year across models, cast to int.
- For text fields (change_summary, latest_year_geography_description):
  - Produce a concise "best consensus" rewrite that preserves only what multiple annotators agree on.
  - Do NOT introduce new details that appear in only one model.
- Always output a single coherent final JSON.

D. When to force human review
Set needs_human_review=true if any of the following holds:
- Major contradictions in change direction or land-use type across models.
- Large numerical disagreement: any change-related score has max gap >= 3 across models.
- latest_year differs by >= 4 years across models (suggesting a reference-year mistake).
- The outputs look generic/template-like with little grounding and inconsistent details.

Output requirements:
- Your entire response must be a single JSON object matching the provided schema.
- No extra text outside JSON.
\end{verbatim}
\end{tcolorbox}
\end{minipage}
\end{figure*}

\end{document}